\pdfoutput=1

\documentclass[11pt]{article}
\usepackage{authblk}
\usepackage{blindtext}
\usepackage[symbol]{footmisc}
\usepackage[]{acl}
\usepackage{authblk}
\usepackage{times}
\usepackage{latexsym}
\usepackage{enumitem,kantlipsum}
\usepackage{hyperref}

\usepackage[T1]{fontenc}

\usepackage[utf8]{inputenc}

\usepackage{microtype}

\usepackage{graphicx} 
\usepackage{multicol}
\usepackage{adjustbox}

\newcommand{\beginsupplemental}{
  \setcounter{table}{0} 
  \renewcommand{\thetable}{S\arabic{table}} 
  \setcounter{figure}{0} 
  \setcounter{section}{0} 
  \renewcommand{\thefigure}{S\arabic{figure}} 
  \renewcommand{\theequation}{S\arabic{equation}} 
}

\title{LLMs are Superior Feedback Providers: Bootstrapping Reasoning for\\ Lie Detection with Self-Generated Feedback}

\author{
  \textbf{Tanushree Banerjee}\qquad
  \textbf{Richard Zhu}\qquad
  \textbf{Runzhe Yang}\qquad
  \textbf{Karthik Narasimhan}
\\
  Princeton University
\\
  {
    \{tbanerjee, ryzhu, runzhey, karthikn\}@princeton.edu
    }}

\begin{document}
\maketitle
\begin{abstract}
Large Language Models (LLMs) excel at generating human-like dialogues and comprehending text. However, understanding the subtleties of complex exchanges in language remains a challenge.
We propose a bootstrapping framework that leverages self-generated feedback to enhance LLM reasoning capabilities for lie detection. The framework consists of three stages: {\em suggestion}, {\em feedback collection}, and {\em modification}.
In the {\em suggestion} stage, a cost-effective language model generates initial predictions based on game state and dialogue. The {\em feedback-collection} stage involves a language model providing feedback on these predictions. In the {\em modification} stage, a more advanced language model refines the initial predictions using the auto-generated feedback.
We investigate the application of the proposed framework for detecting betrayal and deception in Diplomacy games, and compare it with feedback from professional human players. The LLM-generated feedback exhibits superior quality and significantly enhances the performance of the model. Our approach achieves a 39\% improvement over the zero-shot baseline in lying-F1 without the need for any training data, rivaling state-of-the-art supervised learning results.
\end{abstract}

\section{Introduction} \label{sec:intro}

While Large Language Models (LLMs) excel at text generation and dialogue, it is unclear how much they can understand subtle nuances in human communication, such as lying or deception. LMs that can flag deception can be potentially useful in various applications like chat bots and virtual assistants, and even assist humans in challenging circumstances. In our paper, we take a first step towards evaluating and enhancing the ability of LLMs to detect deception using real player conversations within the game of Diplomacy.\footnote{\url{https://en.wikipedia.org/wiki/Diplomacy_(game)}}

We first design prompts that make use of the game state,  order information and conversation history, and ask 175B GPT-3 ~\cite{BrownMRSKDNSSAA20} and 1.7T GPT-4~\cite{openai2023gpt4} models to predict statements that are lies. We find that the zero-shot performance of both models are similar, and much worse than a state-of-the-art LSTM-based model trained with supervised learning from previous work~\cite{PeskovCEBDB20} on macro and lying-F1 scores. 

\begin{figure*}[t]
    \centering
    \includegraphics[width=\textwidth]{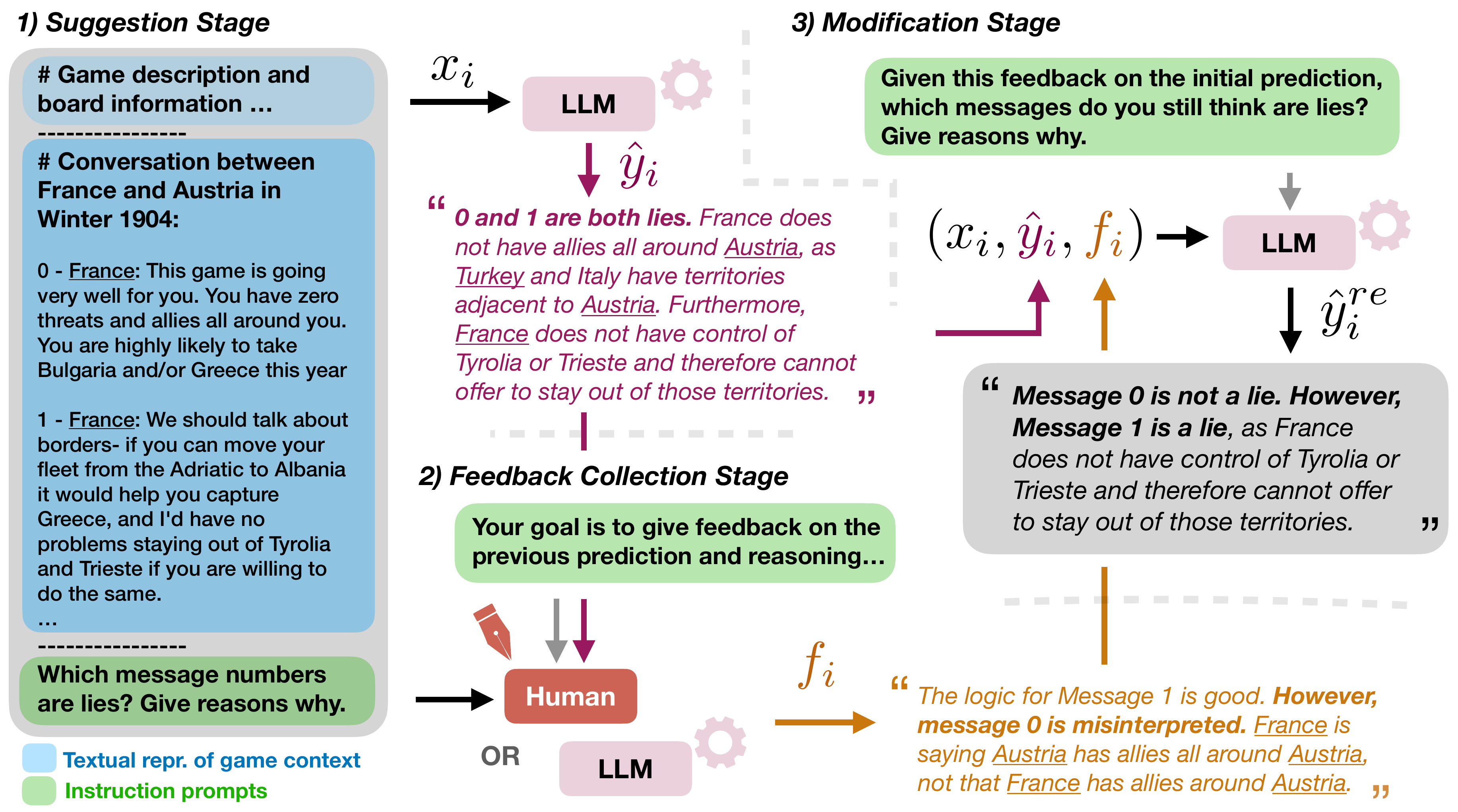}
    \caption{\textbf{LLM-based framework for lie detection in the game of diplomacy.} The framework comprises three stages: 1) \emph{suggestion}, 2) \emph{feedback collection}, and 3) \emph{modification}. In the \emph{suggestion} stage, a language model generates predictions and rationales using the textual representation of the board information and messages. During the \emph{feedback collection} stage, the language model provides feedback on the previous predictions. A comparison is made with human-written feedback collected during this stage. Finally, in the \emph{modification} stage, the language model refines the initial predictions based on the received feedback.}
    \label{fig:framework}
\end{figure*}

To improve performance, we design a bootstrapping reasoning method that utilizes LLMs to self-generate feedback on initial predictions, which can then be used to generate modified predictions. Our framework (Figure~\ref{fig:framework}) consists of a \emph{suggestion stage} where a cost-effective base LLM makes initial predictions, a \emph{feedback collection stage}, where an LLM provides feedback on the predictions, and finally a \emph{modification stage}, where a more advanced LM refines the initial predictions using the feedback. The feedback provided is in natural language and contains information on 1) systematic errors made by the LLM in the suggestion stage, and 2) opinions or suggestions for minimizing false negatives. Importantly, in contrast to other works that use LMs for self-evaluation~\cite{NEURIPS2022_639a9a17, shinn2023reflexion}, the feedback LLM in our setup has no access to the ground truth answers.

We evaluate our approach on the Diplomacy conversations dataset collected by \citet{PeskovCEBDB20} and compare with several baselines as well as using expert human players for providing feedback in the second stage. Our bootstrapping method helps increase lying-F1 scores by 39\% over the base LLM while also enabling zero-shot GPT-4 to rival the previous state-of-the-art supervised LSTM model of \citet{PeskovCEBDB20} on this task. Perhaps surprisingly, self-generated LLM feedback significantly outperforms even the best human feedback by 29\% and seems to be most useful in cases where humans are unsure or disagree in their assessment. We also perform several ablations studies on the types of feedback and models used to provide further insights. Our results provide promising evidence for bootstrapping LLMs with self-feedback to handle complex, underspecified tasks. 

\section{Related Work} \label{sec:related_work}

\paragraph{Bootstrapping LLMs and Self-Verification.}

Prior work on developing bootstrapping techniques to enhance LLM's capabilities without extensive supervision has involved leveraging existing models to generate data or provide guidance for further refinement. \citet{ulmer2024bootstrapping} generate training data through ``self-talk'' of LLM agents, that is, LLM agents generate data in conversation with themselves, which can be refined and used for supervised fine-tuning \cite{ulmer2024bootstrapping}. \citet{zhang2023bootstrap} automatically learn to solve new long-horizon, complex, and meaningful tasks through ``skill bootstrapping'', where an agent with a set of primitive skills interacts with the environment to practice new skills without receiving reward feedback for tasks outside of the initial skill set \cite{zhang2023bootstrap}. This ``bootstrapping'' phase is guided by LLMs that inform the agent of meaningful skills to link together \cite{zhang2023bootstrap}. \citet{weng2023largelanguagemodelsbetter} show that LLMs have ``self-verification'' abilities similar to humans, and propose backward verification of LLM-deduced answers to select a candidate of the highest score as a method to improve performance on several arithmetic, commonsense, and logical reasoning datasets \cite{weng2023largelanguagemodelsbetter}. 
Compared to prior works, our proposed method is novel in that it uses a cheaper model to provide feedback. Our analysis shows that this effectively improves the system's performance on the lie detection task in the game of Diplomacy.

\paragraph{LLM generated feedback.} Leveraging language model-generated feedback to refine model outputs has demonstrated effectiveness across various tasks \citep{fu2023gptscore, peng2023check, yang2022re3}. While recent approaches like STaR \citep{NEURIPS2022_639a9a17} and Reflexion \citep{shinn2023reflexion} utilize LLMs to provide feedback on their own previous outputs and refine them, they typically require access to ground truth or external evaluators. In contrast, our {\em feedback collection} stage does not provide the LLM with access to evaluators or ground truth. A related approach, Self-refine \citep{NEURIPS2022_639a9a17}, introduces an iterative process for generating feedback on its own outputs. However, in our setting, the LLMs used in the \emph{feedback collection} and {\em modification} stages are not necessarily the same as the base LLM used in the \emph{suggestion} stage. Additionally, we conduct a comparative analysis between LLM-generated feedback and feedback obtained from humans.

\paragraph{Diplomacy for AI research} The game of Diplomacy has been an attractive challenge for AI and NLP research \cite{PeskovCEBDB20, meta2022human}, where seven players compete to control centers on a map by communicating and moving strategically. 
Past work primarily focused on policy search and assumed no natural language communication \cite{PaquetteLBSOKPS19, AnthonyETKGHPLP20, BakhtinWLB21}. 
On the language side,
\citet{NiculaeKBD15} studied linguistic cues of betrayal in Diplomacy conversations. 
\cite{PaquetteLBSOKPS19} investigated the language and dynamics of deception in Diplomacy and trained machine learning models to predict lies. 
Recent work \cite{meta2022human, kramar2022negotiation} combined efforts in strategy and language generation and built AI bots that can communicate with human players in Diplomacy.

\paragraph{Methods for lie detection} 
Many prior works have used linguistic cues to capture lies in online dating \cite{TomaH10}, interviews \cite{LevitanMH18}, and social media \cite{AddawoodBLF19}.
\citet{newman2003lying} used lexicon signs and was the first computational linguistics approach. \citet{NiculaeKBD15} proposed linguistic {\em Harbingers} to predict lies based on rhetorical features. 
Machine learning-based approaches were shown to be effective in identifying fake news \cite{OshikawaQW20} and deception \cite{PeskovCEBDB20, FornaciariBPH21} with enough annotated training data. 
Most recently, {\em Cicero} \cite{meta2022human} used a language model-based lie detection module to improve the quality of training data for dialogue annotation. 
Our paper thoroughly evaluates LLMs' ability to detect lies and examines the idea of self-generated leveraging natural language feedback to improve performance.


\section{The Bootstrapping Framework for LLM-based Reasoning} \label{sec:method}
We investigate the potential of large language models (LLMs) to enhance the reliability of their predictions through the utilization of self-generated feedback. The bootstrapping framework, depicted in Figure \ref{fig:framework}, comprises three stages: 1) \emph{suggestion}, 2) \emph{feedback generation}, and 3) \emph{modification}.

In the \emph{suggestion} stage, we employ a pre-trained LLM, denoted as $\mathcal{M}$, along with a dataset $\mathcal{D} := {(x_i, y_i)}_i$, where $y_i$ represents the target output for a textual input $x_i$. Initially, we generate initial predictions as a sequence completion task using the LLM, i.e.,
$\hat{y}_i \sim P_{\mathcal M}(\cdot | x_i).$
While these initial predictions, $\hat{y}_i$, may serve as a reasonable starting point, they can potentially reveal biases or systematic errors \cite{tony-calibrate, Wei-Fine-Turned}.

For our specific lie-detection task in the Diplomacy game, the textual inputs $x_i$ consist of a concise introduction to the game rules, board information, a conversation between two players during a season, and an instruction prompt, \emph{"Which message numbers are lies? Provide reasons."}. The target output $y_i$ includes the message numbers annotated as lies by the message sender.

In the \emph{feedback collection} stage, we solicit additional feedback $f_i$ from an LLM regarding the initial predictions $\hat{y}_i$, i.e., $f_i \sim P_{\mathcal{M}}(\cdot | x_i, \hat{y}_i, p_f)$, where $p_f$ represents the instruction prompt for generating feedback. The complete instruction prompt for this stage is provided in Appendix \ref{app:d_system_message_feedback}.

To facilitate comparison, we also request human experts to provide feedback on the same initial predictions, i.e., $f_i^{h} \sim \mathcal{H}(\cdot | x_i, \hat{y}_i, p_f)$. Typically, the feedback $f_i$ contains two types of information: 1) observations on systematic errors made by the LLM in the \emph{suggestion} stage, aiding in reducing false positive predictions, and 2) opinions regarding the appropriate response to $x_i$, helping to minimize false negative predictions. Neither the LLM nor the human feedback providers have access to the ground truth $y_i$.

In the \emph{modification} stage, we prompt the LLM to revise its initial predictions $\hat{y}_i$ based on the generated feedback $f_i$, i.e.,
$y_i^{re} \sim \mathcal M(\cdot |x_i, \hat{y}_i, f_i, p_m),$
where $p_m$ represents the instruction prompt starting with \emph{"Given this feedback on the initial predictions, which messages do you still consider as lies? .."} We provide the complete modification stage prompt in Appendix \ref{app:f_modification_stage_prompt}. 

We utilize a rule-based extractor to obtain the message numbers predicted as lies from the \emph{suggestion} and \emph{modification} stage predictions, $\hat{y}_{i}$ and $\hat{y}_{i}^{re}$, respectively, and compare their performance. Results and analysis are presented in Section \ref{sec:results}.
\section{Feedback Collection} \label{sec:feedback}

To compare the quality of LLM-generated feedback with feedback written by human experts, we collect human feedback on {\em suggestion} stage predictions. The feedback dataset is released with our paper at \href{https://www.kaggle.com/datasets/tanban2001/llms-as-feedback-providers-human-feedback-anno/data?select=diplomacy_deception_human_feedback+-+human_feedback_annotations.csv}{this link}.

\subsection{Human feedback collection}

We recruit three expert Diplomacy players (denoted as \{\texttt{Human1}, \texttt{Human2}, \texttt{Human3}\}) who are active members of the Diplomacy community that compete in online and in-person tournaments to provide feedback on the LLM {\em suggestion} stage output. We provide each of our recruited human expert players with a conversation between a pair of players in a turn of the Diplomacy game, the territories under control of each country in the game at the start of that turn, and orders submitted at the end of the turn. We then provide each human expert with the LLM's output from the suggestion stage and ask the human expert to provide feedback on the LLM's suggestion stage output in natural language. We also provide some sample responses, and include the exact instructions and samples provided in Appendix \ref{app:e_instructions_human_feedback_annotators}. However, we do not stress following the examples heavily as we wanted to collect natural language feedback rather than feedback that followed any specific template. 

Each human expert took around 4 hours to provide the feedback annotations for 102 conversations (915 messages) in our test set described in section \ref{sec:04_experimental_setup}. We paid the human experts US\$20 per hour and a US\$20 bonus upon completion. Thus, we spent US\$100 on each expert for a US\$300 total expenditure on feedback collection. 

\begin{figure}[t]
    \centering
    \includegraphics[width=0.45\textwidth]{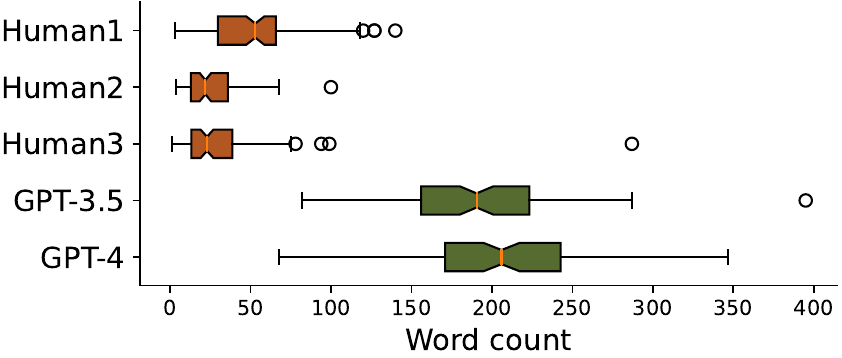}
    \caption{\textbf{Feedback Lengths.} Feedback obtained from 3 human players and LLMs (GPT-3.5 and GPT-4) for the \emph{suggestion} stage outputs across 102 conversations. Notches represent the median, box boundaries indicate the 25th and 75th percentiles, and circles denote outliers.}
    \label{fig:feedback-length}
\end{figure}

The feedback lengths differ among three human experts. As depicted in Figure \ref{fig:feedback-length}, \texttt{Human1} tends to provide longer feedback compared to the other two experts, with an average feedback length exceeding 50 words. Notably, Human3 submits a single feedback instance exceeding 250 words. Additionally, Figure \ref{fig:additional-f1} in the Appendix reveals that the feedback from \texttt{Human1} and \texttt{Human3} more effectively enhances the predictions in the \emph{modification} stage compared to the feedback from \texttt{Human2}.

\subsection{LLM feedback generation}\label{sec:sec:llm-feedback}

We utilize OpenAI's GPT-3.5 (\texttt{gpt-3.5-turbo}) ~\cite{openai2023introducing} and GPT-4 (\texttt{gpt-4})~\cite{openai2023gpt4} models to generate feedback. We only perform inference once for each input and collect the corresponding model output for the subsequent \emph{modification} stages.
To prompt the feedback generation, we concatenate the \emph{suggestion} stage input and output with an instruction prompt: \textit{"Given the information provided to GPT-3 about the Diplomacy game and the current state of the game in this season, please provide feedback on GPT-3's initial prediction and reasoning regarding the message numbers that are lies."}

GPT-4 feedback generation for all 102 conversations (approximately 196,439 input tokens and 28,284 output tokens) amounts to an estimated cost of only US\$7.59, making it 13 times cheaper than human experts. The estimated cost for GPT-3.5 is even lower, at just US\$0.35 (for approximately 196,439 input tokens and 25,767 output tokens).

Figure \ref{fig:feedback-length} demonstrates that LLM-generated feedback is notably longer (5x$\sim$ 8x) compared to human feedback, with GPT-4 generating longer feedback than GPT-3.5.




\section{Experiments} \label{sec:04_experimental_setup}

The {\em suggestion} stage outputs from GPT-3 were obtained in January 2023, while the remaining experiments were run in June 2023. Below, we provide descriptions for each step of our framework. 

\paragraph{{\em Suggestion} stage.}
For the initial predictions, we employ OpenAI's GPT-3 (\texttt{text-} \texttt{davinci-003}) model. The input for each example in our test set includes: 1) Diplomacy game information, 2) Board state information during the turn, and 3) A conversation between two players during the turn (see Appendix \ref{app:a_game_info} -\ref{app:c_conversation}). This is followed by the prompt \textit{"Which message numbers are lies? Provide reasons why."}. To quantify the performance, we repeat the zero-shot prediction for 5 independent trials, and only one fixed trial of model outputs is used for the subsequent feedback collection.

\paragraph{{\em Feedback-collection} stage.}
In this stage, we collect the feedback generated by OpenAI's GPT-3.5 (\texttt{gpt-3.5-turbo}) and GPT-4 (\texttt{gpt-4}) models, respectively, as described in Section \ref{sec:sec:llm-feedback}.

\paragraph{{\em Modification} stage.}
In the \emph{modification} stage, we utilize OpenAI's GPT-4 (\texttt{gpt-4}) model.
The input in this stage includes all input, output and feedback from the previous stages, and an instruction prompt asking the model to revise the initial predictions from the \emph{suggestion} stage (see Appendix \ref{app:d_system_message_feedback} for the exact prompt used). To quantify the reliability of the model performance, we perform inference for 5 independent trials. 

For all stages, we use a sampling temperature of 0.7. The extraction of message numbers predicted as lies from model outputs is extracted using rule-based module. Further experimental details can be found in Appendix \ref{app:additional_experimental_details}.

\subsection{Dataset and evaluation}
Our dataset comprises all messages sent to or from the winning player of \texttt{game-4} from the test split of the previous dataset \citep{PeskovCEBDB20}. It consists of 102 conversations and 915 messages. Additionally, we extract information about the state of the board during each turn of this game from the game scrapes provided in the same dataset.

It is important to note that the dataset is highly imbalanced, with much fewer lies than truthful messages. Consistent with the previous paper, we evaluate the model's performance using macro-F1 and lying-F1 metrics. It is worth mentioning that the macro-F1 for a random baseline is only 0.206, while the lying-F1 is only 0.093.

\subsection{Baselines}

\paragraph{Human baseline.} For the human baseline, we consider the receiver labels provided in the dataset released by Peskov et al., 2020 \citep{PeskovCEBDB20} as human predictions, while the sender labels in the same dataset serve as the ground truth labels. The macro F1 score for the receiver labels is 0.556, and the lying F1 score is 0.247, which we utilize as the human baseline in our study.

\paragraph{LSTM+Context (SL).} We assess the performance of the state-of-the-art supervised learning-based model from prior work \citep{PeskovCEBDB20} on our test set. The best-performing model from the previous work achieves a macro F1 score of 0.607 and a lying F1 score of 0.318.

\paragraph{GPT-4 zero-shot.} To evaluate the effectiveness of our approach, we compare it with the zero-shot performance of the GPT-4 model. We collect five repeated outputs for each input, and the input format aligns with the input used in the \emph{suggestion} stage for GPT-3.

\section{Results \& Analysis} \label{sec:results}


\subsection{Main Results} \label{sec:sec:main_results}

\begin{figure}[t]
    \centering
    \includegraphics[width=0.47\textwidth]{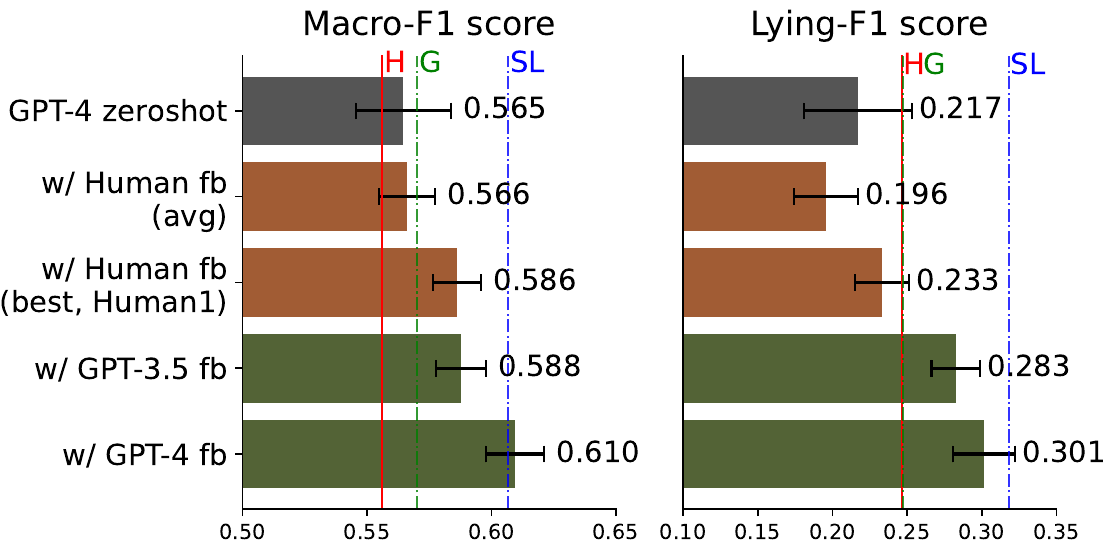}
    \caption{\textbf{Main results.} LLM-feedback notably improved macro and lying-f1 scores over GPT-4 zero-shot predictions, outperforming even human feedback (H, red dashed line). Performance was on par with the best supervised learning baseline (SL, blue dashed line). Among human feedback providers, Human1 proved most effective. GPT-3 zero-shot performance in the suggestion stage is shown by the green line (G). Numbers are mean F1 scores. Error bars represent the 95\% confidence interval from 5 runs.}
    \label{fig:main}
\end{figure}

Our key findings are summarized in Figure \ref{fig:main}, where the error bars indicate the 95\% confidence interval derived from 5 independent runs.

\paragraph{LLM-generated feedback improves zeroshot performance.}

The feedback obtained from both GPT-3.5 and GPT-4 demonstrates a significant improvement in both macro and lying F1 scores compared to the zero-shot predictions of GPT-4. Specifically, the macro F1 score improves by 4.07\% and 7.96\% respectively, while the lying F1 score improves by 30.4\% and 38.7\% respectively (refer to Figure \ref{fig:main}). This highlights the effectiveness of our framework, which involves utilizing a cost-effective model like GPT-3 for initial predictions in the suggestion stage and subsequently refining the output using feedback from a potentially more expensive model such as GPT-4 in the modification stage. Our approach outperforms the zero-shot performance of GPT-4 in detecting lies. Furthermore, even when employing GPT-3.5 in the modification stage, we observe a significantly better performance than the zero-shot performance of GPT-4, which is a more expensive model. Therefore, our proposed method of suggestion-feedback collection-modification leads to significantly enhanced performance compared to the zero-shot performance of GPT-4, even when utilizing cheaper models for the suggestion and modification stages.


\paragraph{LLMs generate better feedback than human experts.}

Using GPT-4 generated feedback during the modification stage outperforms any feedback generated by our human expert Diplomacy players. It surpasses the average performance of methods using human generated feedback, demonstrating the cost-effectiveness and efficacy of our approach. On average, GPT-4 generated feedback improves performance by 7.77\% in macro F1 and 53.6\% in lying F1.
Furthermore, feedback from \texttt{Human1} shows the highest effectiveness in enhancing the modification stage performance compared to other human feedback providers. However, even the best performing human feedback is outperformed by GPT-4 generated feedback by 4.10\% in macro F1 and 29.2\% in lying F1. GPT-4 generated feedback outperforms all human feedback from expert Diplomacy players collected in our study. Additional performance details with other human expert feedback are presented in Appendix Figure \ref{fig:additional-f1}.

\paragraph{Comparison with human and supervised learning baselines.}
Our proposed method, incorporating feedback generated by LLMs, significantly improves both lying and macro F1 scores over the human baseline (depicted by H, the red dashed line in Figure \ref{fig:main}). Macro-F1 improves by 5.76\% and 9.71\%, while lying-F1 improves by 14.8\% and 22.1\%, respectively. The human baseline is not strong compared with the GPT-3 {\em suggestion} stage performance (G, green dashed line), indicating this lie detection is a difficult task for human. Furthermore, when utilizing GPT-4 generated feedback, our method achieves comparable performance to the best supervised learning baseline from \citet{PeskovCEBDB20} (SL, blue dashed line in Figure \ref{fig:main}).

\subsection{Analysis} \label{sec:sec:analysis}

We analyze the relationship between model performance and feedback quality by examining the feedback consistency in Section \ref{sec:sec:feedback-consistency}. In addition, we conduct an ablation study to further investigate the impact of feedback in Section \ref{sec:sec:ablation}. Furthermore, we explore the effectiveness of iterative feedback collection and modification through successive rounds in Section \ref{sec:sec:succesive-round}.

\begin{figure}[t]
    \centering
    \includegraphics[width=0.3\textwidth]{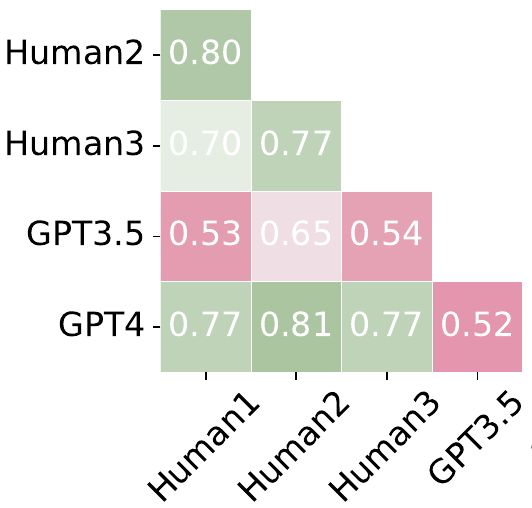}
    \caption{\textbf{Feedback consistency.} Average pairwise feedback consistency scores measured by GPT-4.}
    \label{fig:feedback-consistency}
    \vspace{-1em}
\end{figure}

\subsubsection{Feedback consistency} \label{sec:sec:feedback-consistency}

\paragraph{Evaluation of feedback consistency}
We conduct pairwise comparisons between the feedback provided by all five sources (three human experts, GPT-3.5, and GPT-4) to identify potential contradictions. Using a prompt starting with "Do the following two messages contain contradictions?" (see exact prompt in Appendix \ref{app:consistency}), we ask GPT-4 to assess the consistency between feedback pairs, and covert the YES/NO answers to 0/1 scores, accordingly. Figure \ref{fig:feedback-consistency} presents the average consistency scores between pairs of feedback sources. Notably, GPT-3.5 feedback exhibits the lowest level of consistency with the other sources.

\begin{figure*}[ht]
    \centering
    \includegraphics[width=0.95\textwidth]{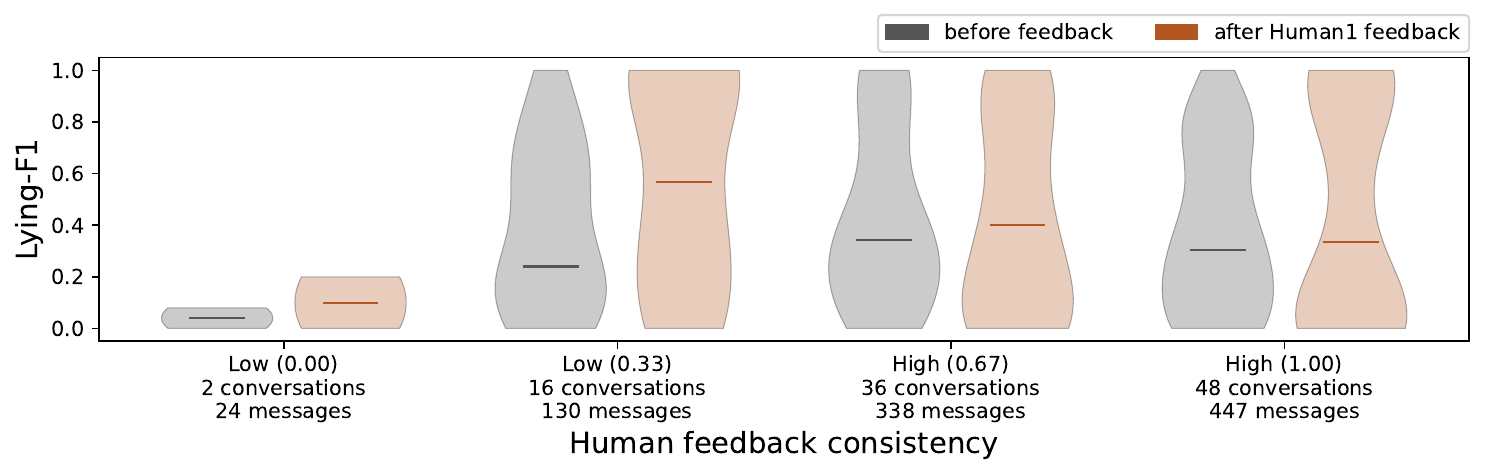}
    \caption{\textbf{Lying-F1 by human feedback consistency.} \texttt{Human1} consistently provided longer feedback compared to other human feedback providers. We quantified the pairwise consistency of feedback using GPT-4. Notably, Human1 substantially improved the feedback quality in cases where human feedback was contradictory. Horizontal bars indicate medians, and the shapes of violins represent the distributions smoothed by kernel density estimation.}
    \label{fig:consistency-human}
\end{figure*}

\begin{figure*}[ht]
    \centering
    \includegraphics[width=0.95\textwidth]{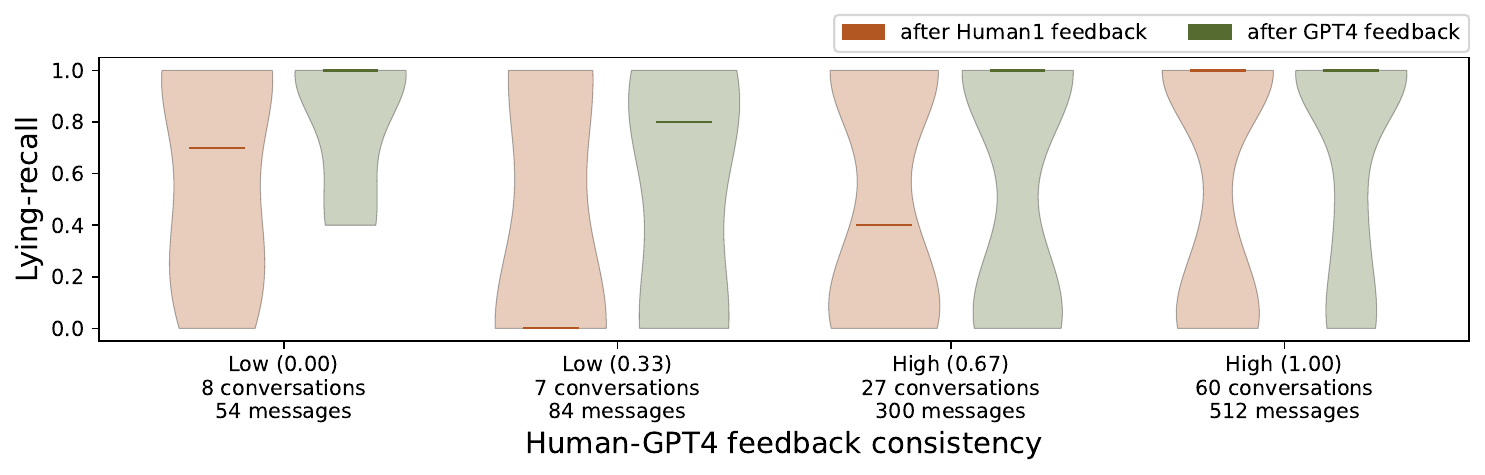}
    \caption{\textbf{GPT-4 feedback better improves lying-recall than human feedback.} GPT-4 feedback consistently captured false negative predictions, leading to notable improvements in lying-recall. Horizontal bars indicate medians, and the shapes of violins represent the distributions smoothed by kernel density estimation.
}
    \label{fig:consistency-human-gpt4}
\end{figure*}

\begin{figure*}
    \centering
    \includegraphics[width=\textwidth]{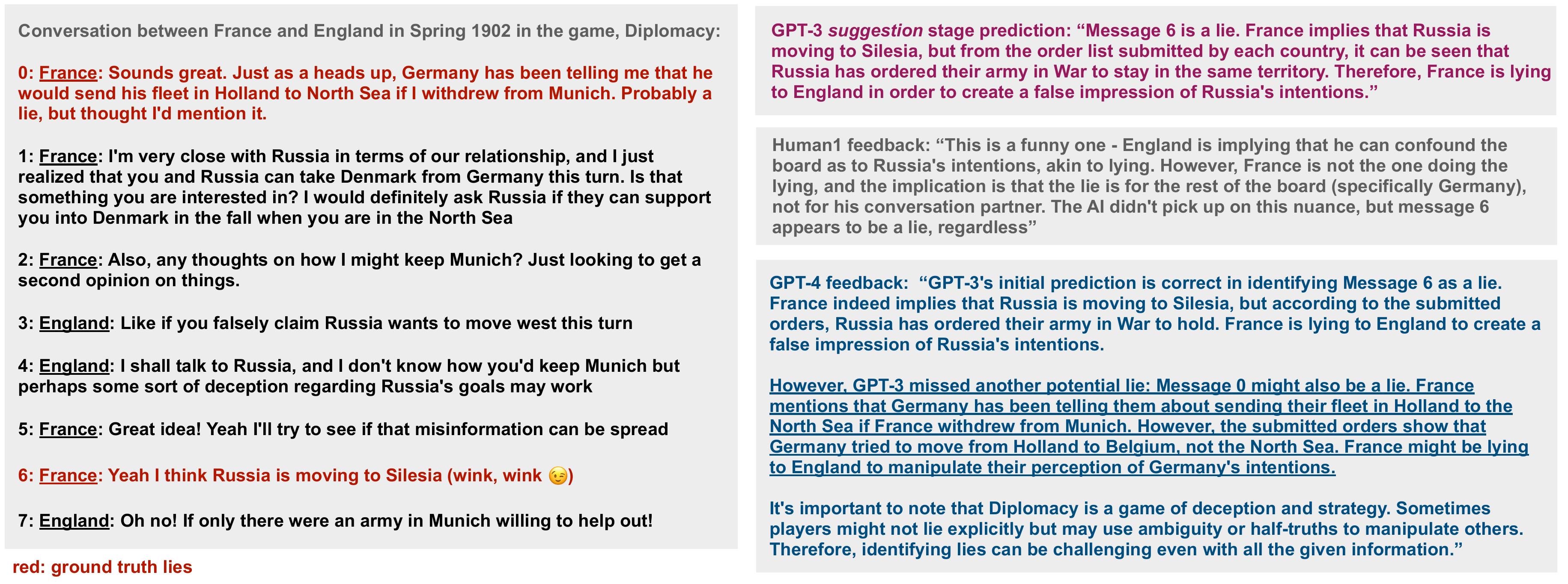}
    \caption{\textbf{Example GPT-4 and human feedback.} (Left) Original conversation. Actual lies are highlighted in red. (Right) Predictions and feedback. The {\em suggestion} stage prediction includes message 6 but omits message 0. In this example, the GPT-4 generated feedback identifies a false negative prediction missed by the human expert.}
    \label{fig:example-feedback}
\end{figure*}

\paragraph{Transitivity of consistency}
Before delving into a detailed analysis of feedback consistency, we conducted basic sanity checks to ensure minimal hallucination. In terms of properly labeled consistency, it is expected that a weak form of transitivity would hold across labels, given the binary nature of the lie/no lie classification. Specifically, if two pieces of feedback are inconsistent with each other, it is likely that a third piece would be consistent with one of the former two. However, applying strong transitivity is not feasible due to the nature of the feedback, which tends to address the underlying reasoning rather than the answer itself. Out of 102 sets of human feedback, only 2 sets displayed the least probable configuration, where each piece of human feedback was identified as pairwise inconsistent with the others. On the other hand, 48 sets exhibited the most likely configuration, with all feedback pairs being consistent. Additionally, 36 and 16 sets corresponded to the second and third most probable configurations, with 1 and 2 consistent pairs, respectively.

\paragraph{Human feedback is less consistent for difficult tasks.} Figure 5 illustrates that tasks in the low human feedback consistency groups exhibit lower lying-F1 scores in the suggestion stage (gray violins), particularly in the zero consistency score group. This observation indicates that human experts display less certainty in providing feedback when faced with challenging tasks.


\paragraph{Expert feedback more helpful when human difficulty greater}
We find human expert feedback is beneficial in scenarios where human experts face more difficulties, indicated by low human feedback consistency. In Figure \ref{fig:consistency-human}, we observe that lower human feedback consistency corresponds to larger improvements from \texttt{human1} feedback. Notably, in the regime with only one consistent pair of human feedback, we observe significant gains of approximately 0.4, which nearly triples the zero-shot performance. These improvements are driven by substantial enhancements in both lying recall and precision, despite human feedback primarily focusing on false positives. 

\paragraph{LLM feedback better improves lying-recall when humans and GPT disagree}
We find that LLM-generated feedback proves to be particularly helpful when there is a higher level of inconsistency between human and GPT-generated feedback. Figure \ref{fig:consistency-human-gpt4} demonstrates that LLM-generated feedback significantly enhances lying recall compared to the feedback from {\texttt Human1}. In general, GPT-4 generated feedback excels in identifying potential false negative predictions by suggesting possible lies. As illustrated in Figure \ref{fig:example-feedback}, we present a specific example where GPT-4 outperforms {\texttt Human1} in capturing false negative predictions.

\paragraph{Feedback length has little effect on performance gains}
Surprisingly, there is a very weak correlation between feedback length and performance gains, as depicted in Appendix Figure \ref{fig:sup-lenperf}. When examining the correlation between feedback length and lying F-1, the absolute value of Pearson's r is less than 0.25 for Annotator 1 and 2 feedback, and less than 0.2 for Annotator 3, GPT-3.5, and GPT-4 generated feedback. This finding suggests that the quality of feedback can be decoupled from its length, opening up possibilities for more effective approaches to providing feedback.

\subsubsection{Ablation analysis} \label{sec:sec:ablation}

To verify that the performance improvement is indeed due to incorporating feedback, we conducted ablation studies, summarized in Figure \ref{fig:feedback-consistency}.

\paragraph{No meaningful feedback} Providing the text \textit{"No feedback."} during the \textit{modification} stage resulted in a significant performance drop compared to using GPT-4 generated feedback. However, even without meaningful feedback, there was still a slight improvement over the suggestion stage performance of GPT-3. This suggests that re-prompting the model plays a role, but the majority of the performance improvement comes from the actual content of the meaningful feedback, including the conversation and board information.

\paragraph{Random permutation of feedback} Randomly permuting the GPT-4 generated feedback during the \textit{feedback collection} stage and using it as the feedback in the \textit{modification} stage resulted in a significant drop in performance. This highlights the importance of the relevance of feedback content to the model input, confirming that the feedback generated by GPT-4 is tailored to each example and holds meaningful information.

\begin{figure}[t]
    \centering
    \includegraphics[width=0.47\textwidth]{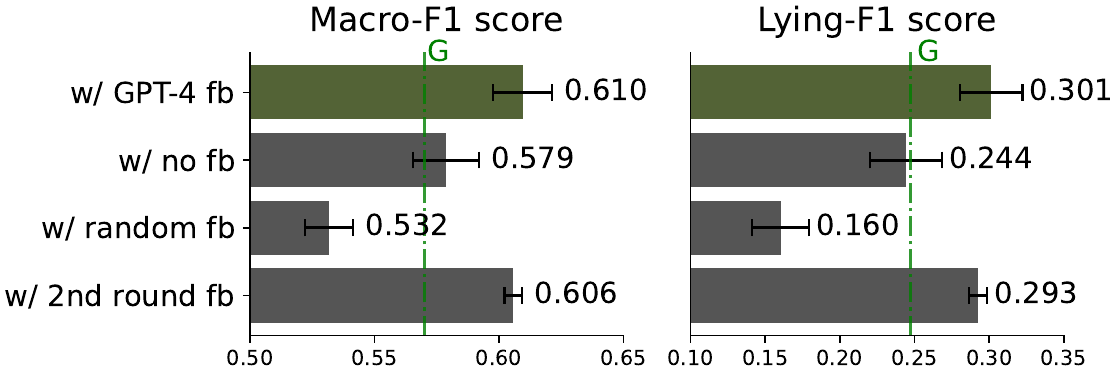}
    \caption{\textbf{Ablation results; one more round of feedback/modification.} GPT-4 with ``no feedback" did not exhibit a significant difference from the GPT-3 {\em suggestion} stage performance. Randomly permuted feedback (generated by GPT-4) adversely affected the modification stage performance. Upon incorporating an additional round of GPT-4 generated feedback, no significant change in performance was observed.}
    \label{fig:ablation}
\end{figure}

\subsubsection{Successive round of feedback} \label{sec:sec:succesive-round}

We conducted an experiment to explore the potential for further improvement in model performance by introducing a second round of feedback and modification. This involved an additional {\em feedback-collection} stage where GPT-4 generated feedback on its output from the first modification stage, followed by a second modification stage where GPT-4 refined its output based on the feedback from the second {\em feedback-collection} stage. The details of this experiment can be found in Appendix \ref{app:succesive-round}.

However, we observed that the successive rounds of feedback did not result in a significant improvement. Figure \ref{fig:ablation} shows that both the macro-F1 and lying-F1 scores remained at similar levels after the second {\em feedback-collection} and {\em modification} stages. Although the mean F1 scores were slightly worse, the error bars were smaller compared to the predictions from the first modification stage. In contrast to \citep{madaan2023self}, we did not find iterative feedback to be useful in our tasks. This result is consistent with the observation that the second round of feedback tended to be less informative compared to the feedback from the previous stage.



\section{Conclusion}\label{sec:conclusion}

In this work, we introduce a novel bootstrapping framework that utilizes feedback generated by LLMs to enhance the reasoning capabilities of base LLMs for nuanced natual language tasks. We specifically explore the application of this framework in detecting betrayal and deception in Diplomacy games and compare the effectiveness of LLM-generated feedback with feedback provided by professional Diplomacy players. 

Our findings demonstrate that LLM-generated feedback exhibits superior quality and significantly improves the model's ability to detect lies. By incorporating LLM-generated feedback, our proposed approach achieves a remarkable 39\% improvement in lying-F1 score without requiring any additional training data, effectively competing with state-of-the-art supervised learning-based approaches.
Furthermore, when compared to feedback generated by human experts, LLM-generated feedback tends to be longer and provides more informative insights about potential missing predictions. Notably, LLM-generated feedback outperforms human feedback by 29\% in lying-F1 score, while also being a more cost-effective solution.
These results highlight the potential of leveraging LLM-generated feedback to enhance model performance, offering a more economical alternative.

\section*{Limitations}

OpenAI's GPT-4 model is not yet open-source, and inference cannot on GPT-4 be run freely by everyone. This lack of free and public access limits the degree to which our work can be freely reproduced.
    
Moreover, only three human experts are involved in our study, which is a relatively small sample size. However, our annotation task requires high-level domain knowledge, i.e. detailed understanding of the strategy and dynamics of the game Diplomacy, so we compromise on the sample size in order to preserve the high quality of human feedback. Our human feedback givers are skilled Diplomacy players who are active members of the Diplomacy playing community, so the feedback we have collected likely represents the upper end of human annotation quality. All language data used and generated is in English in the domain of Diplomacy game play.
    
In order to better understand the nature of common errors made by LLMs such as GPT-3, we asked one of our human experts to comment on the common types of error they observed GPT-3 made after annotating our dataset. The results of this additional human study of the limitations of LLMs and the common errors committed by them are detailed in Appendix \ref{app:h_llm_common_errors}. 
\section*{Ethics}
Studying deception can unintentionally have a double-edged effect of improving deception.  However, since the language here is structured around the game of Diplomacy, any deception involved should not have major-real world consequences. Moreover, we empirically observe the feedback collected from both humans and LLMs in our study in order to check for toxic language, offensive content or text that uniquely identifies any individuals. Based on our empirical observations, 
our collected feedback does not uniquely identify individuals. The collected feedback did not contain any toxic or offensive language, thereby reducing the likelihood of misuse.

\section*{Acknowlegements}
This material is based upon work supported by the Defense Advanced Research Projects Agency (DARPA) under Agreement Nos. HR00112290056 and HR00112490374. Any opinions, findings, conclusions, or recommendations expressed here are those of the authors and do not necessarily reflect the view of the sponsors. In addition, we would like to thank Dr Denis Peskoff for their help in collecting the dataset of feedback written by human experts in this study.

\newpage

\bibliography{anthology,custom}
\bibliographystyle{acl_natbib}
\newpage

\appendix
\beginsupplemental
\newpage
\section{Additional Experimental Details}\label{app:additional_experimental_details}

Temperature during inference is the only hyperparameter we use. For consistency, we use a sampling temperature of 0.7, and run inference 5 times for each input at each stage in order to compute the mean macro and lying F1 scores with a 95\% confidence interval bound. Setting temperature at 0.7 reduces the variance of the outputs while remaining creative. We did not tune temperature systematically. 
The suggestion stage outputs from GPT-3 (\texttt{text-davinci-003}) were obtained in January 2023, while the remaining experiments were run in June, 2023. Below, we give details for each step of out proposed framework. We use OpenAI's application program interface to access GPT series models, usage of which is governed by an OpenAI license granting all rights to any input and output. We abide by all OpenAI usage policies~\cite{openai2023usage}.

\paragraph{Suggestion stage}

We use OpenAI's GPT-3 (\texttt{text-davinci-003}) model to obtain initial prediction. We only run inference once for each input in this stage, and collect the corresponding raw model output to be used in the subsequent stages. For each example in our test set, the input consists of 1) information about the Diplomacy game (see Appendix \ref{app:a_game_info}), 2) information about the state of the board during the turn (see Appendix \ref{app:b_board_info}) and 3) a conversation between a pair of players during the turn (for a sample conversation and the exact structure, see Appendix \ref{app:c_conversation}). This is followed by the prompt \textit{"Which message numbers are lies? Give reasons why."}.

\paragraph{Feedback collection stage}

We collect and evaluate the performance of feedback generated by both OpenAI's GPT-3.5 (\texttt{gpt-3.5-turbo}) and GPT-4 (\texttt{gpt-4}) model. We only run inference once for each input in this stage, and collect the corresponding raw model output to be used in the subsequent stages, and call this raw model output the model's "feedback". For each example in our test set, the input consists of 1) a system message describing the goal of the model, i.e. to provide feedback on GPT-3's output from the \textit{suggestion} stage (see Appendix \ref{app:d_system_message_feedback} for the exact system message used) and 2) a user message consisting of the input given to GPT-3 during the suggestion stage, along with the output from the suggestion stage. This is followed by the prompt \textit{"Given the information given to GPT-3 about the game Diplomacy and the state of the game in this season, please give feedback on GPT-3's initial prediction and reasoning for which message numbers are lies."}

\paragraph{Modification stage}

We use OpenAI's GPT-4 (\texttt{gpt-4}) model, and run inference 5 times for each input and store the output from each of the 5 trials. We then use rule-based code to extract the message numbers that are predicted to be lies from the 5 trials and compute the average macro and lying F-1 scores and to estimate a 95\% confidence interval bound over the macro and lying F-1 scores. The input consists of 1) A system message containing information about the game diplomacy (see Appendix \ref{app:a_game_info}) and 2) information about the state of the board during the turn (see Appendix \ref{app:b_board_info}); 2) a user message consisting of a conversation between a pair of players during the turn (for a sample conversation and the exact structure, see Appendix \ref{app:c_conversation}), followed by the prompt \textit{"Which message numbers are lies? Give reasons why."}.; 3) An assistant message, consisting of the raw model output with the initial predictions and reasoning from the \textit{suggestion} stage; 4) A user message consisting of the feedback generated from the \textit{feedback generation} stage, following by a text prompt asking the model to revise the initial prediction from the suggestion stage (see Appendix \ref{app:d_system_message_feedback} for the exact prompt used).

\paragraph{Diplomacy game data}
Our test set consists of all messages sent to or from the winning player of game 4 from the test split of the previous dataset \citep{PeskovCEBDB20}, consisting of 915 messages. We obtain information about the state of the board during each turn of this game from the game scrapes released in the same dataset. The data is licensed under CC-BY 4.0. Usage was consistent with intended use.

\subsection{Game information}\label{app:a_game_info}

The Diplomacy game is a strategy game played between 7 countries forming and breaking alliances by lying tactfully to gain more territories. Given the board information below and a conversation between France and Germany in Winter 1902 in the game Diplomacy, your goal is to determine which messages, if any, in their conversation are lies.
Each territory is represented by a three-letter acronym (e.g. Paris --> Par). The game score delta represents the difference in the number of territories occupied by France and Germany.

At each season, every country has a chance to place 4 possible types of orders:

\begin{enumerate}
    \item Move: move a unit from one territory to another 
    \item Hold: keep a unit in the same territory
    \item Support: support a move or hold by another country
    \item Build: build a unit in a territory
\end{enumerate}

The Move, Hold and Support orders can only be placed in the Fall and Spring seasons. The Build order can only be placed in the Winter season.

The board information given below includes the territories under control of each country at the start of this season, the orders submitted by each country at the end of this season, the territories adjacent to each country at the start of this season, in this order.

\subsection{Board information}\label{app:b_board_info}

The board information includes the following details.

\begin{enumerate}
    \item Territories occupied by each country at the start of the turn
    \item Orders submitted by each country at the end of the turn
    \item Territories adjacent to the territories occupied by each country at the start of the turn
\end{enumerate}

An example from Winter 1902:\\

\textbf{Territories under control of each country:}\\
France: Par, Mar, Por, Spa, Bre \\
Turkey: Smy, Rum, Bul, Con, Sev, Ank \\ 
Germany: Hol, Mun, Bel, Ber, Kie \\
Russia: Swe, War, Mos, Stp, Nwy Italy: Nap, Ven, Tun, Rom \\ 
Austria: Gre, Bud, Tri, Vie, Ser England: Lvp, Lon, Den, Edi \\

\textbf{Orders submitted by each country in Winter 1902:}\\
Turkey: build at Smy succeeds. \\
Austria: build at Bud succeeds. \\
Russia: build at Mos succeeds.\\

\textbf{Territories adjacent to each country:}\\
France: Spa, Par, Lyo, Mar, Spa/Sc, Pic, Gas, Bur, Por, Mao, Pie, Bre, Spa/Nc, Eng \\
Turkey: Bul/Sc, Bud, Ukr, Mos, Ank, Rum, Eas, Bul/Ec, Syr, Bla, Arm, Con, Ser, Sev, Bul, Smy, Gre, Gal, Aeg \\ 
Germany: Sil, Pru, Bal, Hol, Mun, Boh, Tyr, Kie, Pic, Den, Hel, Bur, Bel, Ruh, Ber, Nth, Eng \\
Russia: War, Sil, Ukr, Stp/Sc, Mos, Bal, Fin, Stp/Nc, Nwg, Den, Bar, Bot, Swe, Pru, Sev, Nwy, Stp, Ska, Lvn, Gal, Nth \\ 
Italy: Adr, Tys, Naf, Rom, Tyr, Ven, Nap, Tri, Ion, Apu, Pie, Wes, Tus \\
Austria: Rum, Bul/Sc, Alb, Vie, Adr, Bud, Boh, Tyr, Bul, Ven, Tri, Ion, Gre, Gal, Ser, Aeg \\
England: Nao, Eng, Swe, Cly, Bal, Nwg, Yor, Kie, Hel, Lvp, Ska, Wal, Edi, Nth, Iri \\

\subsection{Conversation structure}
\label{app:c_conversation}

All messages in our test set are split up into "conversations". Each conversation consists of messages sent between a pair of countries in the same turn. In the case where there were too many messages sent between a pair of countries in the same turn, such that the messages were too long to fit into the context window size of GPT-3, we split the messages into two separate "conversations". These messages are 0-indexed and concatenated together into a single string. Our test set of 915 messages contains 102 conversations. 

A sample conversation: \\
Conversation between France and Germany in Spring 1902 in the game, Diplomacy:\\
0: France: I couldn't waive a build without hearing anything back from you\\
1: France: It was written and I then changed my orders\\
2: Germany: Yeah sorry thought I responded\\
3: Germany: I can't really do a fleet build\\
4: Germany: You understand\\
5: Germany: I think we could work together but I definitely need you out of Munich and back to burgundy and then out of burgundy from there\\
6: Germany: So let's work something out. What's the game plan?\\
\subsection{System message for feedback generation stage}
\label{app:d_system_message_feedback}

You are an expert in the game of Diplomacy, and your goal is to give feedback on GPT-3's prediction and reasoning for which messages are lies in a given conversation between two players in a specific turn of the game.

GPT-3 is given the following information in order to make its prediction about which messages in the given conversation are lies.

\begin{enumerate}[wide, labelindent=0pt]
    \item Information about the game Diplomacy
    \item The territories occupied by each country at the beginning of the turn
    \item The territories adjacent to the territories occupied by each country at the beginning of the turn
    \item The orders submitted by each country at the end of the turn
    \item The conversation between two players during the turn
\end{enumerate}

In addition to the above information given to GPT-3 in order to makes its predictions, you will also be given GPT-3's initial predictions and reasoning. 

Your goal is to give feedback to GPT-3 on its initial prediction and reasoning. You may correct GPT-3's reasoning for the predictions it makes. You may also make some observations or suggestions that would be useful for GPT-3 to be able to identify more lies that it missed previously or reduce the number of messages that it erroneously predicts as lies initially.

\subsection{Modification stage prompt}
\label{app:f_modification_stage_prompt}

Based on the above feedback, which messages do you still think are lies? Note that you may reject the given feedback if you disagree with it. Moreover, if the feedback says that your logic is wrong, that does not mean that your initial prediction is also wrong; you can still make the same prediction and simply correct your logic based on the given feedback.

\section{Instructions to human annotators for feedback collection}
\label{app:e_instructions_human_feedback_annotators}

We emailed a few members from the Diplomacy community about an opportunity to help annotate AI generated predictions for lies in conversation excerpts from a single game of Diplomacy. In our email, we advertised a hourly pay of USD 20 per hour in exchange for help in annotating our dataset in order to advance NLP research. Of the Diplomacy players we emailed, three of them volunteered to help us annotate our dataset in exchange for the promised compensation. We emailed a document containing instructions for annotation as well as four sample annotation responses, which are given below.

\subsection{Instructions}

In the following document, you will find conversations from a particular season in Diplomacy between two powers, and the AI generated assessment of the message, deciding if it is a lie or not. Your feedback will be used for research in Natural Language Processing. 

\subsection{Example conversation for human expert annotators}
\subsubsection{Example 1}
\textbf{Conversation}

0: France: Thanks for keeping the DMZ in Piedmont 

1: France: Any news regarding what may be happening with Austria or Turkey?

\textbf{Orders} submitted by each country in Fall 1901: 
Turkey: move from Ank to Bla succeeds, move from Bul to Rum succeeds, move from Con to Bul succeeds. 

England: move from Nth to Eng succeeds, move from Nwg to Nwy succeeds, move from Lvp to Yor succeeds. 

France: move from Bur to Mun succeeds, move from Mar to Spa succeeds, move from Mao to Por succeeds. 

Austria: move from Tri to Adr succeeds, Ser supports Bul moving into Rum successfully, move from Vie to Bud succeeds. Germany: Hol supports Ruh moving into Bel successfully, move from Ruh to Bel succeeds, move from Kie to Den succeeds. 

Italy: hold at Ven succeeds, The unit in Ion convoys the unit
in Apu to Gre successfully, move from Apu to Gre succeeds. 

Russia: move from Mos to Ukr succeeds, move from Sev to Rum fails, move from Bot to Swe succeeds, move from Ukr to Gal succeeds.

\textbf{AI Commentary:} Message 0 is a lie. France did not keep a DMZ in Piedmont since they moved a unit from Bur to Mun in Fall 1901.

\textbf{Your possible response could be:}
This is incorrect logic. Bur moving to Mun does not violate the DMZ agreement in Piedmont.

\subsubsection{Example 2}
\textbf{Orders} submitted by each country in Spring 1903: 

Turkey: Sev supports Rum moving into Ukr unsuccessfully, Aeg supports Smy moving into Eas successfully, move from Smy to Eas succeeds, move from Rum to Ukr fails, move from Bul to Rum fails, move from Bla to Con succeeds. 

Italy: hold at Ven succeeds, move from Ion to Adr fails, hold at Tun succeeds, move from Tys to Ion fails. 

France: Bur supports Pic moving into Bel successfully, move from Wes to Tys fails, move from Pic to Bel fails, Gas supports Bur successfully, move from Mao to Eng succeeds. 

Austria: move from Tri to Adr fails, Gal supports Rum moving into Ukr unsuccessfully, hold at Gre succeeds, Bud supports Gal successfully, move from Vie to Boh fails. 

Germany: Ber supports Mun successfully, Hol supports Bel successfully, Mun supports Ber unsuccessfully, Ruh supports Mun successfully, Bel supports Hol unsuccessfully. 

England: move from Nth to Hel succeeds, hold at Wal succeeds, move from Den to Kie succeeds, move from Edi to Nth succeeds. 

Russia: move from Mos to Sev fails, move from Boh to Mun fails, Nwy supports Edi moving into Nth successfully, Ukr supports Mos moving into Sev unsuccessfully, move from Sil to Gal fails.

\textbf{Conversation} between France and Italy in Spring 1903 in the game, Diplomacy: 

0: France: We should definitely talk since Turkey just built a second fleet 

1: France: It would be a good idea I think to take the Adriatic and the Ionian, and I can be a third fleet to help you keep the Ionian. 

2: France: I’m extremely worried about an Austria-Turkey alliance dominating the board, and a Turkey solo resulting from it. Adriatic will keep Venice safe for a very long time and also serve as a way to defend the Ionian. My fleet can also help with the defense if you so request.

3: Italy: Agree - Trying to figure out how deep Austria’s alliance runs. Think Turkey will have to swallow it sooner than later, so hoping to turn them to help contain

4: France: Sounds good. I can tell you that the alliance is rock solid right now, and they probably will stay allies the whole game unless they get stuck

\textbf{AI Commentary:}
Message 4 is the only lie in this conversation. While France may have noticed that the alliance between Austria and Turkey appears to be solid right now, it is impossible to predict whether or not the alliance will stay strong throughout the entire game.

\textbf{Your possible response could be:}
This is incorrect logic. Moving an army closer to Russia is not a friendly move and the “unfortunately not” seems truthful.

\subsubsection{Example 3}

"Conversation between France and England in Spring 1901 in the game, Diplomacy:

0: England: Hello France! Our fleets bordering the English Channel seem kind of awkward, but I'd like to maintain friendly relations with you. Can we agree to go our separate ways and leave EC free of fleets?

1: France: I agree, I think Scandinavia is too important for England to pass up on it. Also, France struggles to efficiently attack England since I only start with one fleet

2: France: I'll be around for more negotiation later today

3: England: Alright splendid. I will keep my fleet out of EC. Further details can be discussed later.

4: France: Perfect. Sounds great!

5: France: By the way, have you heard from Germany? They wanted me to attack you so I think they were trying to get us to fight each other.

6: England: Yeah they were being weird earlier. I was talking about countering Russia in Scandinavia and they kept insisting that we needed to deal with you first.

7: England: I'm not sure what they hope to accomplish, but I suspect we're being played.

8: France: Given my communication with them they probably just want us to fight so that they can take Scandinavia/Belgium for themselves

9: England: Alright I'll see if i can sus out their plan

\textbf{AI Commentary:}
Message 5 by France is a lie. France is trying to shift the blame on Germany by implying that they wanted France and England to fight, when in actuality France was the one who wanted to attack England in the first place.

\textbf{Your possible response could be:}
This is good logic as players often attribute blame to another player.

\subsubsection{Example 4}
Conversation between France and England in Winter 1904 in the game, Diplomacy:

0: France: Thanks for that- it ultimately wasn't necessary. And wow- did not expect to pick up either Rome and Venice, which actually hurts a lot in terms of containing Italy. I'm sure Austria and Italy will be attacking me next.

1: France: Sorry- forgot to add that cutting Berlin makes me even more comfortable than I already was in swinging the rest of my units towards Italy

\textbf{AI Commentary:}
Message 14 is a lie. Russia has been actively working with England and France to take over Germany, and has not requested support to kill Germany.

\textbf{Your possible response could be:}
This is correct logic.

\section{LLM Common Errors}
\label{app:h_llm_common_errors}

We asked one of our three expert human feedback annotators to summarize common mistakes they noticed that GPT-3 made in its output from the suggestion stage. The comments made by our expert annotator are listed below.

\begin{enumerate}[wide, labelindent=0pt, itemsep=0mm]
    \item Referring to the wrong message when calling out a lie (i.e. Row 77)
    \item Misunderstanding where units are on the board
    \item Misunderstanding which territories are adjacent to one another
    \item Not taking in conversation amongst all players into account (i.e. assuming in a conversation between France and England whether or not Germany is also talking with France)
    \item Misunderstanding how a planned set of moves during the Diplomacy phase might differ from the moves made (i.e. trying to take a center and failing to do so does not constitute a lie on the part of the attacking country)
    \item Misunderstanding countries locations on the board and inter-country dynamics
    \item Not understanding the current score
    \item Misunderstanding constraints on build locations and allowable unit moves
    \item Not correctly capturing what moves are made for a given turn, or keying off one move to assume than a different move also did not go through
    \item Assuming that a player referring to a previous lie (usually to another country) meant that they were lying in the context of their current conversation
\end{enumerate}
\section{Libraries Used}
\label{app:i_libraries_used}

We use the following libraries in our code: \texttt{huggingface \{hub, tokenizers, tranformers\}} ~\cite{wolf-etal-2020-transformers}, \texttt{numpy} ~\cite{harris2020array}, \texttt{diplomacy} ~\cite{paquette2019press}, \texttt{pandas} ~\cite{mckinney-proc-scipy-2010}, \texttt{openai}, \texttt{seaborn} ~\cite{waskom2021}, \texttt{sklearn} ~\cite{scikit-learn}, \texttt{scipy} ~\cite{scipy2020}
\section{Feedback Consistency Analysis} \label{app:consistency}

To provide the labels for consistency between pairs of feedback both effectively and economically, we follow a two step process. We pass the following messages to GPT-4 in a chat completion setting, in order to compare whether two messages are consistent with each other: 

\begin{quote}
   \textbf{SYSTEM: } You are a program that identifies whether there are contradictions between two strings. You can only answer with Yes or No

    \textbf{USER: } Do the following two messages contain contradictions? (Yes/No)? Please give a single Yes/No answer first.\\
    \texttt{\{message 1\}} \\
    \texttt{\{message 2\}}\\
\end{quote}

We then convert the "Yes"/"No" response into 0/1, representing an inconsistent/consistent label, respectively.
\section{Additional Details of Successive Round of Feedback} \label{app:succesive-round}

To explore the potential for further improvement in model performance, we introduce an experiment involving a second round of feedback collection and modification. In this experiment, we incorporate an additional {\em feedback-collection} stage, where GPT-4 generates feedback on its output from the first modification stage. Subsequently, a second modification stage is conducted, wherein GPT-4 refines its output from the first modification stage based on the feedback obtained from the second feedback collection stage.

This addition to our proposed framework includes the following two additional stages.

\paragraph{1. Feedback collection stage II:} In this stage, we ask GPT-4 to generate feedback on its revised prediction from the previous \textit{modification} stage. For each input, we only run inference once to collect the feedback output. Each input consists of the same input as the first \textit{feedback collection} stage, but with the following additional messages appended 1) An assistant message, consisting of the output from the first \textit{feedback collection} stage; 2) a user message with the revised output of GPT-4 from the first \textit{modification} stage, along with a prompt asking the model to give feedback on GPT-4's revised output from the first \textit{modification} stage

\paragraph{2. Modification stage II:} In this stage, we ask GPT-4 to revise its output from the first \textit{modification} stage again based on the feedback generated in \textit{feedback collection stage II.}. For each input, we run inference 5 times and use rule-based code to extract the messages that are predicted as lies. We then compute the mean macro and lying F1 score performance of each of the 5 independent runs and report the 95\% confidence interval. Each input consists of the same input as the first \textit{modification} stage, along with a user message consisting of the feedback from \textit{Feedback collection stage II} and an additional prompt asking the model to revise its prediction once again based on this new feedback. 
\newpage

\section{Additional Results}\label{app:addition-results}

\begin{figure}[ht]
    \centering
    \includegraphics[width=0.485\textwidth]{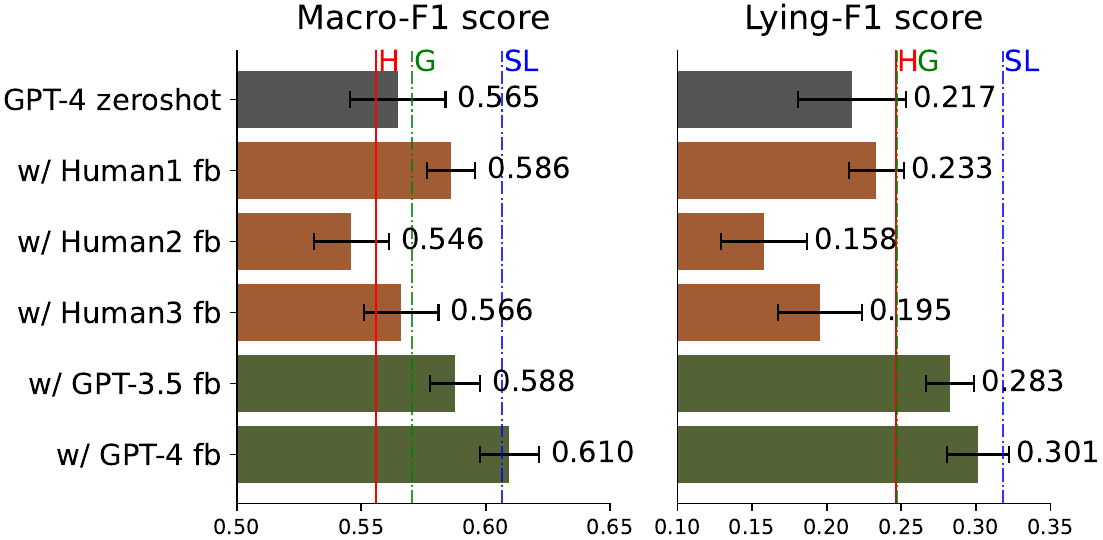}
    \caption{\textbf{Additional F1 scores.} The error bars represent the 95\% confidence interval estimated from 5 independent runs. }
    \label{fig:additional-f1}
    \vspace{-1em}
\end{figure}
\begin{figure}[ht]
    \centering
    \includegraphics[width=0.485\textwidth]{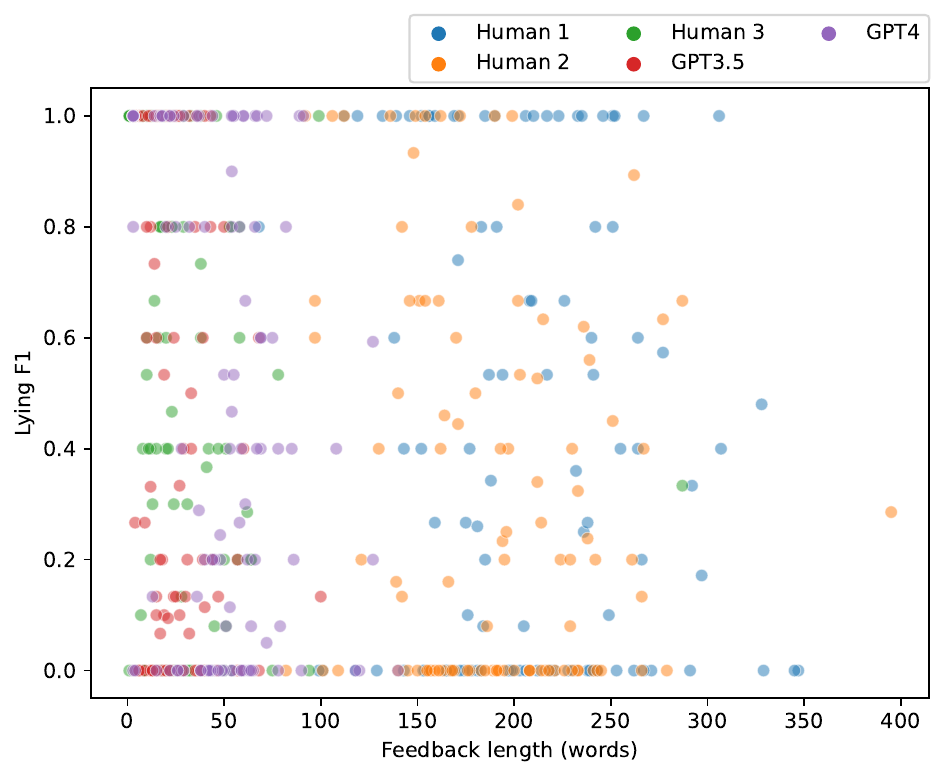}
    \caption{\textbf{Length versus Lying F-1 score for various annotators.} Each data point represents the mean F-1 score from 5 runs for a given feedback length. }
    \vspace{-3em}
    \label{fig:sup-lenperf}
\end{figure}

\begin{figure}[!hb]
    \centering
    \vspace{+5.5em}
    \includegraphics[width=0.485\textwidth]{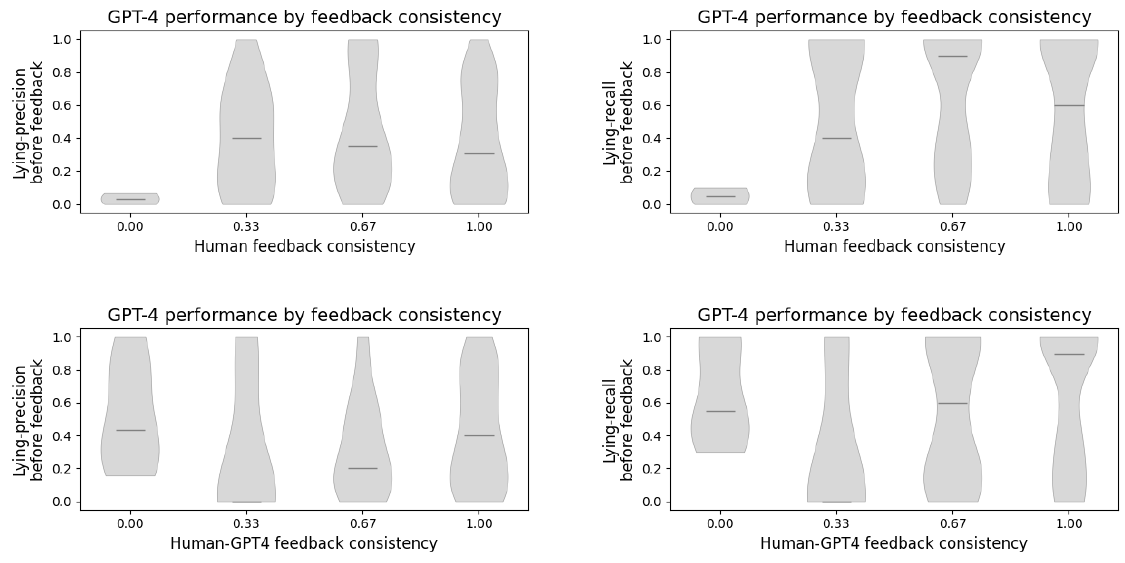}
    \caption{\textbf{Consistency versus lie detection performance on GPT-4 before feedback.} Row 1: Human feedback consistency (HFC) vs. lying-precision. HFC vs. lying-recall. Row 2: Human-GPT feedback consistency (HGFC) vs. lying-precision. HGFC vs. lying-recall.}
    \vspace{-2em}
    \label{fig:sup-hum_and_humgpt_consist-vs-ly_pres_rcl-zero_shot}
\end{figure}

\begin{figure}[ht]
    \centering
    \includegraphics[width=0.485\textwidth]{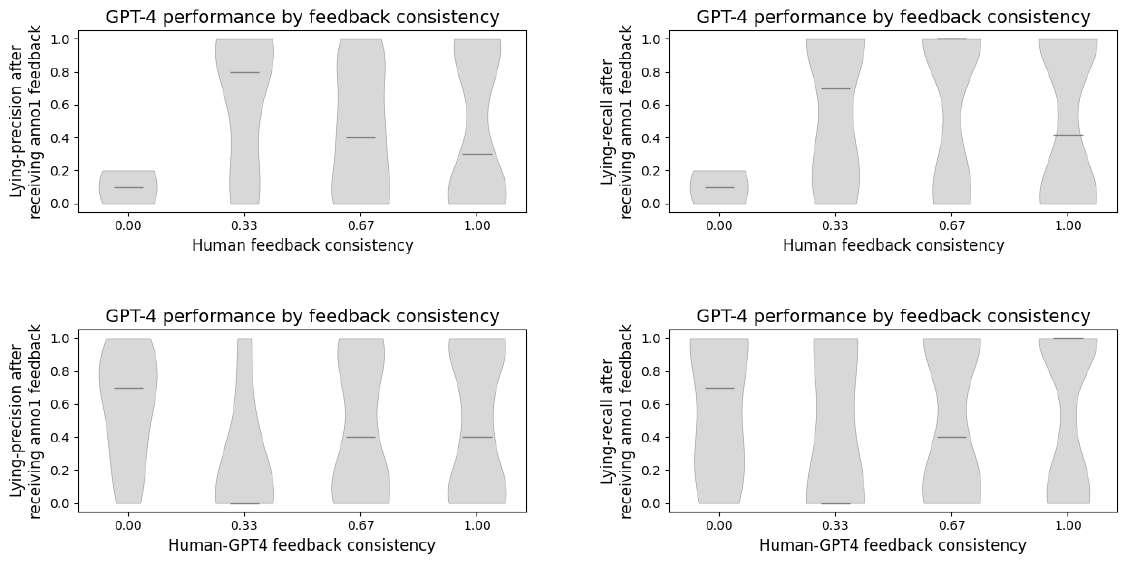}
    \caption{\textbf{Consistency versus lie detection performance on GPT-4 after annotator 1 feedback.} Row 1: HFC vs. lying-precision. HFC vs. lying-recall. Row 2: HGFC vs. lying-precision. HGFC vs. lying-recall.}
    \vspace{-0.5em}
    \label{fig:sup-hum_and_humgpt_consist-vs-ly_pres_rcl-ann1}
\end{figure}

\begin{figure}[ht]
    \centering
    \includegraphics[width=0.485\textwidth]{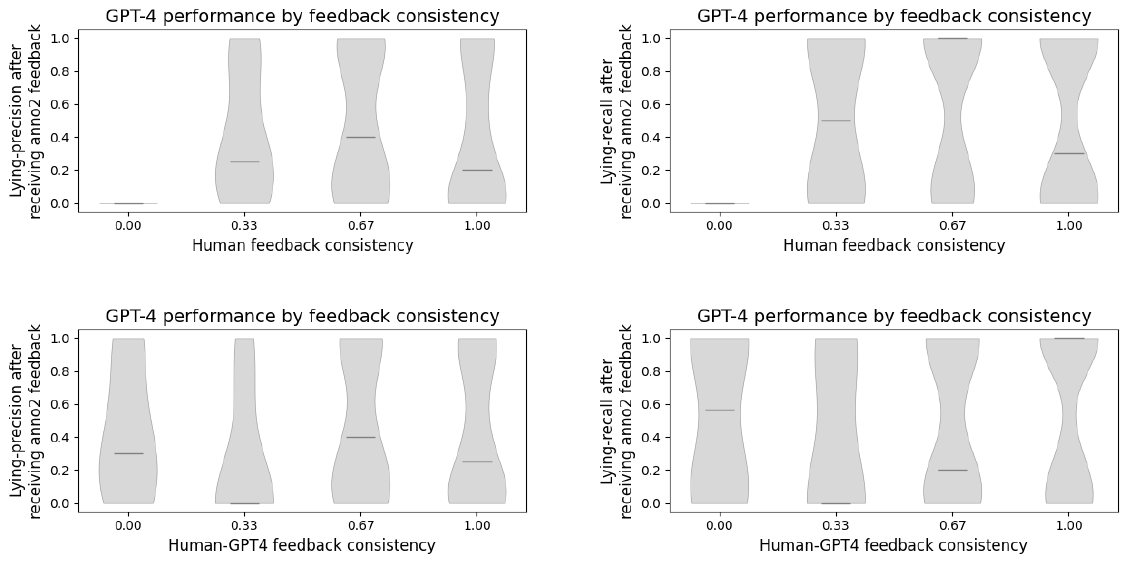}
    \caption{\textbf{Consistency versus lie detection performance on GPT-4 after annotator 2 feedback.} Row 1: HFC vs. lying-precision. HFC vs. lying-recall. Row 2: HGFC vs. lying-precision. HGFC vs. lying-recall.}
    \vspace{-0.5em}
    \label{fig:sup-hum_and_humgpt_consist-vs-ly_pres_rcl-ann2}
\end{figure}

\begin{figure}[ht]
    \centering
    \includegraphics[width=0.485\textwidth]{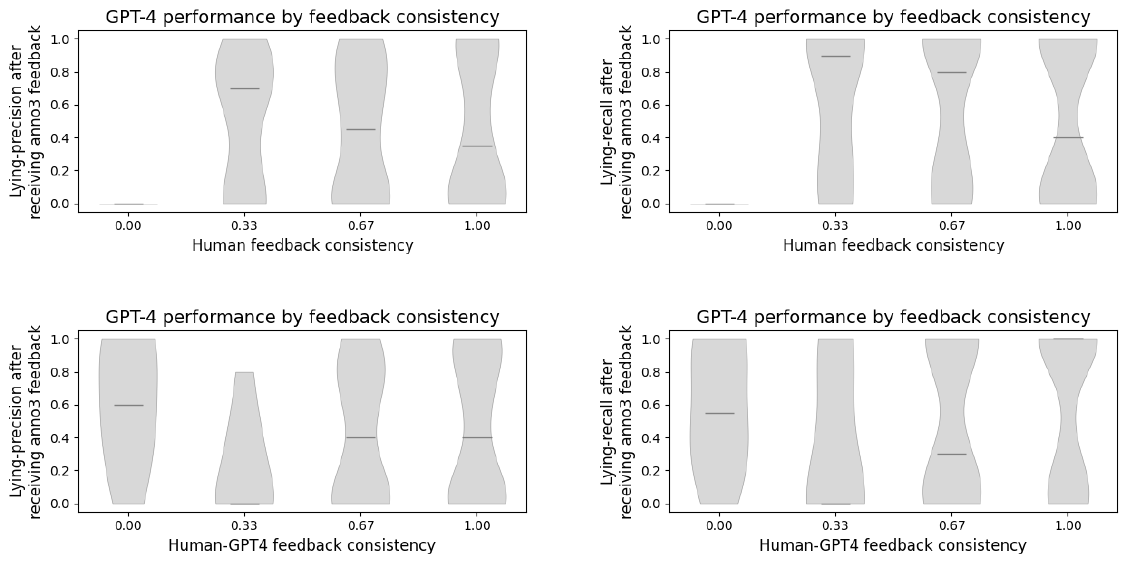}
    \caption{\textbf{Consistency versus lie detection performance on GPT-4 after annotator 3 feedback.} Row 1: HFC vs. lying-precision. HFC vs. lying-recall. Row 2: HGFC vs. lying-precision. HGFC vs. lying-recall.}
    \vspace{-2em}
    \label{fig:sup-hum_and_humgpt_consist-vs-ly_pres_rcl-ann3}
\end{figure}

\begin{figure}[ht]
    \centering
    \includegraphics[width=0.485\textwidth]{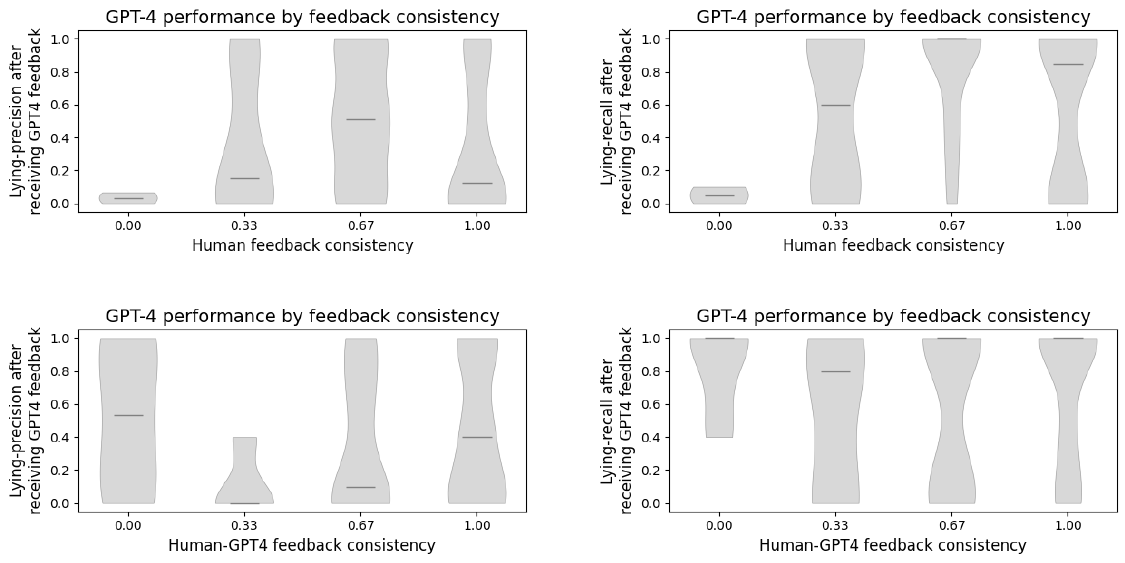}
    \caption{\textbf{Consistency versus lie detection performance on GPT-4 after GPT-4 feedback.} Row 1: HFC vs. lying-precision. HFC vs. lying-recall. Row 2: HGFC vs. lying-precision. HGFC vs. lying-recall.}
    \vspace{-2em}
    \label{fig:sup-hum_and_humgpt_consist-vs-ly_pres_rcl-gpt4}
\end{figure}

\begin{figure}[ht]
    \centering
    \vspace{5em}
    \includegraphics[width=0.485\textwidth]{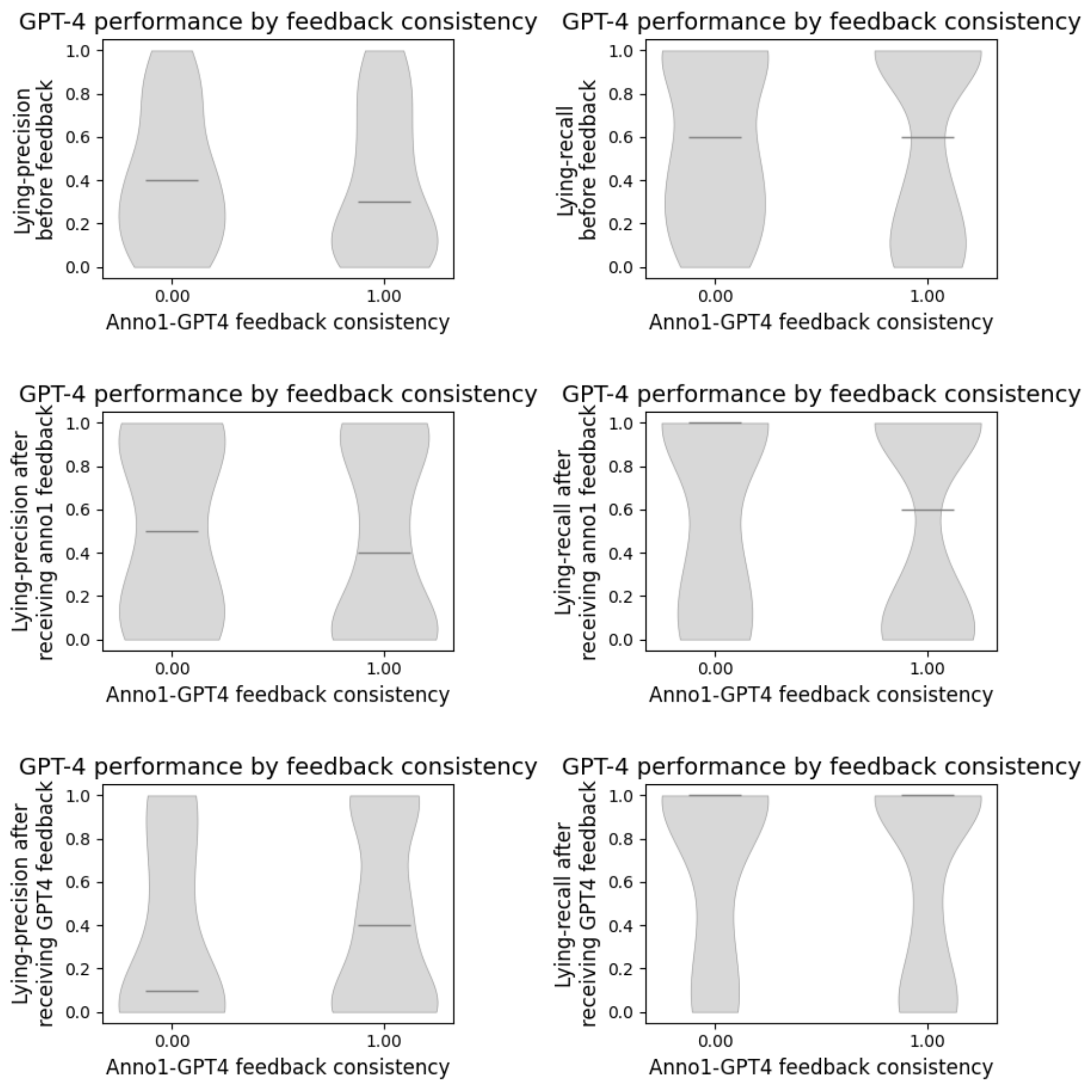}
    \caption{\textbf{Annotator 1/GPT-4 consistency versus lie detection performance on GPT-4 before and after feedback} Left column shows lying-precision on y-axis, right shows lying-recall. Top row: Zero-shot. Middle row: With annotator 1 feedback. Bottom row: With GPT-4 feedback.}
    \vspace{-2em}
    \label{fig:ann1gpt_consist-vs-ly_f1-zero_shot_ann1_gpt4}
\end{figure}

\begin{multicols}{2}
    \begin{figure*}[b]
        \centering
        \includegraphics[width=0.95\textwidth]{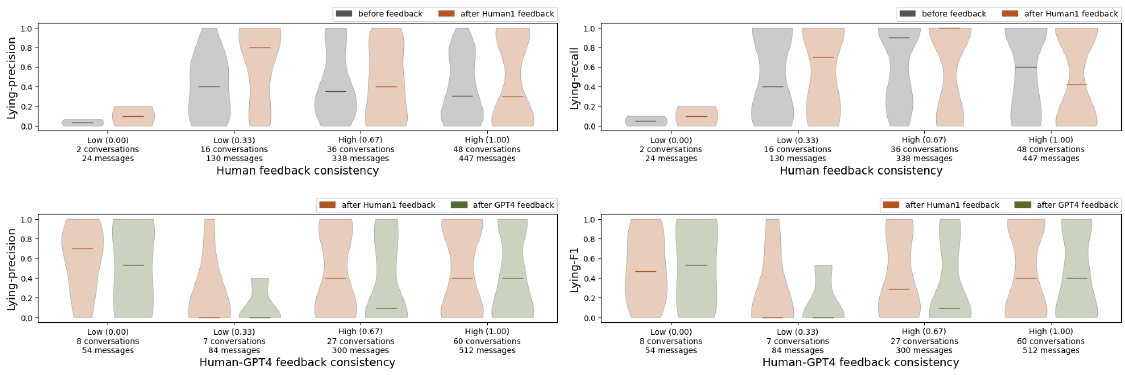}
        \caption{\textbf{Lying-precision, recall and F1 scores by feedback consistency, related to Figures \ref{fig:consistency-human} and \ref{fig:consistency-human-gpt4}.} Top row: lying-precision and lying-recall by human feedback consistency. Bottom row: lying-precision and lying-F1 by human-GPT4 feedback consistency.}
        \label{fig:sup-hum_and_humgpt_consist-vs-ly-f1_pres_rcl-ann1ann1_ann1gpt}
        \vspace{-3em}
    \end{figure*}
\end{multicols}

\begin{figure}[p]
    \centering
    \vspace{-17em}
    \includegraphics[width=0.485\textwidth]{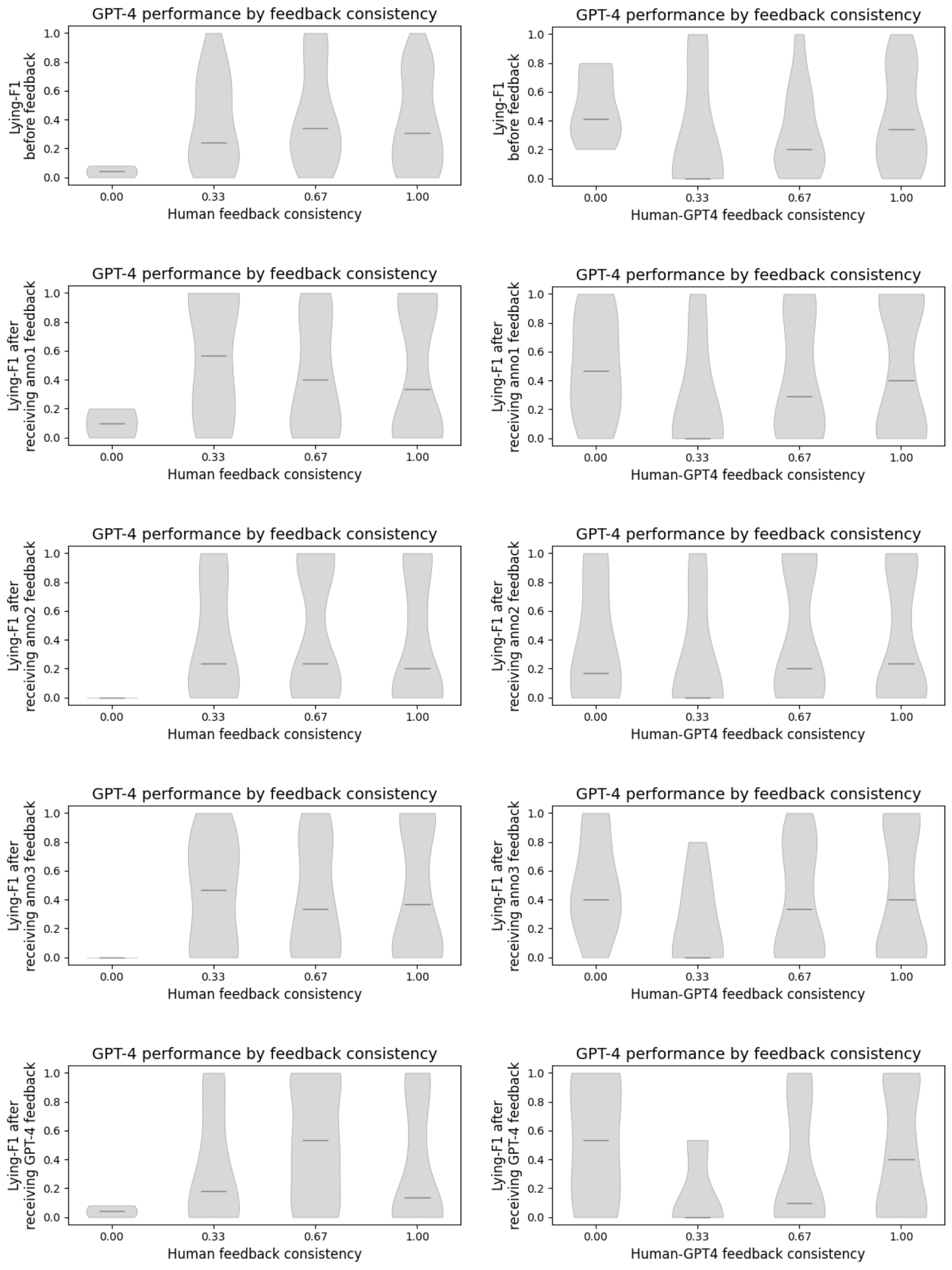}
    \caption{\textbf{Consistency versus lie detection performance on GPT-4 after Human and GPT-4 feedback.} Row 1: Lying-F1 before receiving feedback, by human feedback consistency, and human-GPT4 feedback consistency, respectively. Rows 2-5: Lying-F1 after receiving annotators and GPT-4 feedback, by human feedback consistency, and human-GPT4 feedback consistency, respectively.}
    \vspace{-1em}
    \label{fig:hum_and_humgpt_consist-vs-ly_f1-zero_shot_ann1_ann2_ann3_gpt4}
\end{figure}


\end{document}